\documentclass[conference]{IEEEtran}
\IEEEoverridecommandlockouts
\usepackage{cite}
\usepackage{amsmath,amssymb,amsfonts}
\usepackage{graphicx}
\usepackage{textcomp}
\usepackage{xcolor}
\usepackage{algorithm}
\usepackage{booktabs}
\usepackage{multirow}
\usepackage{amsmath, amssymb}
\usepackage[caption=false,font=footnotesize]{subfig}

\makeatletter
\@ifpackageloaded{algorithmic}{%

}{}
\makeatother

\usepackage{float}
\floatstyle{ruled}
\newfloat{algo}{tbp}{loa}
\floatname{algo}{Algorithm}

\usepackage[noend]{algpseudocode}

\makeatletter
\newcommand{\StateNoNum}[1]{%
  \Statex \hspace*{\dimexpr\ALG@thistlm+\algorithmicindent\relax}#1%
}
\makeatother

\usepackage{iftex}
\ifPDFTeX
  \usepackage{textcomp} 
  
  \newcommand{\equalmarkinline}{\textdagger}
\else
  \newcommand{\equalmarkinline}{⚔}
\fi

\def\BibTeX{{\rm B\kern-.05em{\sc i\kern-.025em b}\kern-.08em
    T\kern-.1667em\lower.7ex\hbox{E}\kern-.125emX}}
\begin{document}

\title{Slim Scheduler: A Runtime-Aware RL and Scheduler System for Efficient CNN Inference\\
}

\author{%
\IEEEauthorblockN{Ian Harshbarger $\dagger$}
\IEEEauthorblockA{\textit{Computer Science}\\
\textit{University of California Irvine}\\
iharshba@uci.edu}
\and
\IEEEauthorblockN{Calvin Chidambaram $\dagger$}
\IEEEauthorblockA{\textit{Computer Science}\\
\textit{Northwood High School}\\
calvinchidambaram@gmail.com}
}

\IEEEaftertitletext{%
  \noindent\centering
  {\normalfont\normalsize \equalmarkinline\ Equal contribution by marked authors.}%
  \par\addvspace{.4\baselineskip}%
}

\maketitle
\begin{abstract}
Most neural network scheduling research focuses on optimizing static, end-to-end models of fixed width, overlooking dynamic approaches that adapt to heterogeneous hardware and fluctuating runtime conditions. We present \textbf{Slim Scheduler}, a hybrid scheduling framework that integrates a Proximal Policy Optimization (PPO) reinforcement learning policy with algorithmic, greedy schedulers to coordinate distributed inference for slimmable models. Each server runs a local greedy scheduler that batches compatible requests and manages instance scaling based on VRAM and utilization constraints, while the PPO router learns global routing policies for device selection, width ratio, and batch configuration. This hierarchical design reduces search space complexity, mitigates overfitting to specific hardware, and balances efficiency and throughput. Compared to a purely randomized task distribution baseline, Slim Scheduler can achieve various accuracy and latency trade-offs such as: A 96.45\% reduction in mean latency and a 97.31\% reduction in energy usage dropping accuracy to the slimmest model available (70.3\%). It can then accomplish an overall reduction in average latency plus energy consumption with an increase in accuracy at the cost of higher standard deviations of said latency and energy, effecting overall task throughput. 



\end{abstract}


\section{Introduction}
Deep neural networks have achieved widespread use across domains such as image classification, object detection, and perception. However, their deployment in real-time and resource-constrained settings (e.g., autonomous systems) remains limited by high inference latency and energy demands. In environments where multiple heterogeneous devices share workload execution, efficient runtime scheduling becomes as critical as model design itself.

Traditional approaches to efficient inference, such as model pruning, quantization, and slimmable networks, reduce computational cost but typically rely on static or heuristic runtime control. These methods seldom exploit cross-device parallelism or dynamic resource feedback, leading to suboptimal utilization under varying loads. Furthermore, most schedulers operate at the level of individual models or layers, ignoring the inter-segment dependencies that emerge when inference is distributed across multiple GPUs.

In this work, we introduce a multi-device, runtime-aware scheduler that distributes segmented neural network inference across heterogeneous GPUs. Our approach combines algorithmic and learning-based strategies through a hybrid PPO+greedy framework. The greedy scheduler manages local batching and execution, grouping compatible requests by segment and slimming width while respecting memory and utilization constraints. It prevents overloading by dispatching only to idle instances and adaptively scales instance counts based on observed demand. Above this layer, a PPO router learns global routing decisions—selecting which device, width ratio, and batch configuration to use based on telemetry of latency, energy, and utilization imbalance.

This hierarchical design enables efficient multi-device coordination without retraining or manual tuning. We implement the system on a segmented SlimResNet backbone, using per-segment slimming to adjust channel widths dynamically. Experiments on CIFAR-100 show that our scheduler substantially reduces average latency and energy consumption compared to baseline greedy and random strategies while maintaining comparable accuracy. These improvements come with a measured trade-off of higher latency and energy variance, reflecting the system’s adaptive exploration of width configurations. The learned PPO policy generalizes across hardware, demonstrating that reinforcement learning can capture device-agnostic scheduling patterns for distributed inference.

\section{Related Works}
Efficient neural network inference has been widely studied through model compression and adaptive computation. Slimmable neural networks \cite{Yu2018, Cai2019} introduced dynamic width scaling, allowing a single model to operate under multiple computational budgets without retraining. These approaches reduce active channels or layers based on runtime constraints, enabling flexible accuracy–latency tradeoffs. Subsequent work, such as SLEXNet \cite{SLEX}, extended this concept by jointly optimizing depth and width scaling with on-device runtime schedulers for embedded systems. While effective on single accelerators, these techniques primarily address intra-model efficiency and do not explicitly coordinate multi-device execution or cross-server resource balancing.

Parallel and distributed inference frameworks have sought to improve throughput by dividing computation across multiple devices. Pipeline-parallel systems such as GPipe \cite{Huang2018} and PipeDream \cite{Narayanan2019} distribute network stages across GPUs to maximize utilization and reduce latency. However, these methods generally assume static execution plans and uniform device capabilities, limiting adaptability to runtime load variations. In contrast, our work focuses on runtime-aware scheduling that dynamically allocates workloads across heterogeneous GPUs with differing performance characteristics.

Dynamic resource scheduling has also been explored in broader computing contexts. Traditional schedulers rely on heuristic rules for load balancing in distributed or cloud systems \cite{Vijay2023, Yu2022}, while recent advances employ reinforcement learning to adapt task placement and resource allocation to time-varying workloads \cite{Li2024}. AutoDistill \cite{Chen2023} applies a similar learning-based approach for distributed inference scheduling but assumes static model architectures. Although these schedulers optimize system throughput and latency under uncertainty, they generally operate at the task level and do not exploit internal neural network flexibility such as tunable width or adaptive batch configuration.

Our approach bridges these domains by integrating slimmable neural architectures with reinforcement learning–based scheduling. Unlike static or single-device adaptation methods, we introduce a PPO-guided multi-GPU scheduler that learns to jointly select model width, batch size, and device assignment. The result is a hybrid framework that combines algorithmic greedy batching with learned global coordination, achieving adaptive load distribution and energy–latency balance across heterogeneous hardware. 

\section{System Design and Methodology}

\begin{algo}[t]
\caption{Greedy Segment–Slim Scheduler (single server)}
\label{alg:greedy}
\begin{algorithmic}[1]
  %
  \StateNoNum{\textbf{Require:} Arrival rate $r$, batch limit $B_{\max}$, VRAM cap $M_{\max}$ (GB), util block threshold $U_{\mathrm{blk}}$ (\%), idle unload $t_{\mathrm{idle}}$, scale trigger $Q_{\mathrm{th}}$, scale cap $N_{\mathrm{new}}$, slimming set $\mathcal{W}$.}
  \StateNoNum{\textbf{State:} FIFO queue $Q$ of $q_t(\mathrm{seg}, w_{\mathrm{req}}, t_{\mathrm{enq}}, \hat w_{\mathrm{prev}})$; loaded instances list $I$ of $i_s(\mathrm{seg}, w, \mathrm{device}, \mathrm{busy}, t_{\mathrm{last}})$; recent GPU util samples $U$.}

  \Procedure{LOOP}{}
    \While{true}
      \State Wait until $Q$ non-empty; peek head key $(\hat s,\hat w,\hat w_{\mathrm{prev}})$
      \State Form batch $B \subseteq Q$ of up to $B_{\max}$ requests with $(\hat s,\hat w,\hat w_{\mathrm{prev}})$
      \State $inst \gets$ \Call{FindFreeBestFit}{$I$, $\hat s$, $\hat w$} \Comment{smallest width $\ge \hat w$}
      \If{$inst=\varnothing$}
        \State $inst \gets$ \Call{CanLoad}{$\hat s,\hat w$}
        \If{$inst=\varnothing$}
          \State Requeue $B$ to front of $Q$; \textbf{continue}
        \EndIf
      \EndIf
      \State $inst.\mathrm{busy}\gets \mathrm{true}$; run \Call{RunBatch}{$inst,B$}; update $inst.t_{\mathrm{last}}$
    \EndWhile
  \EndProcedure

  \Function{FindFreeBestFit}{$I,s,w_{\mathrm{req}}$}
    \State \Return free $i\!\in\!I$ with $i.\mathrm{seg}=s$ and minimal $i.w \ge w_{\mathrm{req}}$ (or $\varnothing$)
  \EndFunction

  \Function{CanLoad}{$s,w$}
    \State Estimate bytes of $(s,w)$; query $(\mathrm{VRAM}_{\mathrm{used}},\mathrm{VRAM}_{\mathrm{tot}})$
    \If{$\mathrm{VRAM}_{\mathrm{used}}+\mathrm{bytes} > M_{\max}$} \State \Return false \EndIf
    \State $u \gets$ latest GPU util from $U$
    \If{$u \neq \varnothing \land u \ge U_{\mathrm{blk}}$} \State \Return false \EndIf
    \State \Return true
  \EndFunction

  \Procedure{UnloaderLoop}{}
    \While{true}
      \ForAll{non-busy $i\in I$}
        \If{$t_{\mathrm{now}} - i.t_{\mathrm{last}} \ge t_{\mathrm{idle}}$}
          \State offload to CPU, free VRAM, remove from$I$
        \EndIf
      \EndFor
    \EndWhile
  \EndProcedure
\end{algorithmic}
\end{algo}

\subsection{Greedy Scheduler (baseline)}
We implement a multi-threaded, best-fit greedy executor for a segmented, universally slimmable backbone. Incoming requests are enqueued with key \(k=(s, w_{\text{req}}, w_{\text{prev}})\), where \(s\) is the segment index and \(w\) is the width (slimming ratio). The worker repeatedly forms a batch from the FIFO head’s key and assigns it to a free instance of the same segment with the smallest width \(w \ge w_{\text{req}}\). If no such instance exists, the scheduler \emph{opportunistically scales up} by instantiating up to \(N_{\text{new}}\) additional instances for key \(k\), guarded by a VRAM budget \(M_{\max}\) and a live GPU-utilization block threshold \(U_{\text{blk}}\). Idle instances are offloaded after \(t_{\text{idle}}\) to release memory. The system samples utilization and emits  telemetry data (utilization, VRAM, per-segment queue sizes, latency percentiles) to support profiling and as input for PPO model training. This greedy executor serves as the local dispatch layer within our PPO\,+\,greedy hybrid: PPO provides high-level routing and overload signals, while the greedy policy delivers low-overhead batching and responsive scale-up under bursty load. Key knobs: \(r, B_{\max}, M_{\max}, U_{\text{blk}}, t_{\text{idle}}, Q_{\text{th}}, N_{\text{new}}, \mathcal{W}\).

\subsection{PPO Router (high-level policy).}

We train a factored PPO router that, given compact telemetry, jointly chooses the target server, model width, and micro-batch group. The server head uses an $\varepsilon$-mixed likelihood (with on-policy correction in the PPO ratio) to encourage exploration across $N$ backends. Rewards couple an accuracy prior (from the 4-stage width tuple) with latency, energy, and a cross-server imbalance penalty, aligning learning with the trends noticed in Fig.1-3. We use one-step advantages with normalization, a clipped surrogate, value loss, and an entropy bonus. The learned router provides high-level dispatch signals that the greedy executor realizes locally (forming batches, scaling instances within VRAM/utilization limits), yielding a practical PPO\,+\,greedy hybrid for multi-server distribution under bursty load.

\paragraph{\textbf{PPO Model Setup}}
At scheduling step $t$, the state vector encodes global and per-server telemetry:
\begin{equation}
    s_t=\big[\,\underbrace{q^{\mathrm{fifo}}_t,\; c^{\mathrm{done}}_t}_{\text{global}},\;\underbrace{\{(q^{(i)}_t,\; P^{(i)}_t,\; U^{(i)}_t)\}_{i=1}^N}_{\text{for each of $N$ servers}}\big],
\end{equation}

where $q^{\mathrm{fifo}}_t$ is the total FIFO length, $c^{\mathrm{done}}_t$ the completed count, and for server $i$: queue length $q^{(i)}_t$, power $P^{(i)}_t$ (J), and GPU utilization $U^{(i)}_t$ (\%). The action is factored into three categorical choices
\begin{equation}
    a_t \triangleq (a^{\mathrm{srv}}_t,\; a^{\mathrm{w}}_t,\; a^{\mathrm{g}}_t),
\end{equation}

selecting the server index $srv$, slimming width $w$, and micro-batch group size $g$. A shared MLP yields logits for each head and a scalar value:
\begin{equation}
    (\ell^{\mathrm{srv}}_\theta,\;\ell^{\mathrm{w}}_\theta,\;\ell^{\mathrm{g}}_\theta,\;V_\theta)(s_t)=\mathrm{MLP}_\theta(s_t).
\end{equation}

Conditioned on $s_t$, the policy factorizes as a product of categoricals,

\begin{equation}
\begin{aligned}
    \pi_\theta(a_t\mid s_t)=\pi^{\mathrm{srv}}_\theta(a^{\mathrm{srv}}_t\mid s_t)\cdot \pi^{\mathrm{w}}_\theta(a^{\mathrm{w}}_t\mid s_t)\cdot \pi^{\mathrm{g}}_\theta(a^{\mathrm{g}}_t\mid s_t), \\ 
    \pi^{\bullet}_\theta(\cdot\mid s_t)=\mathrm{Cat}\!\big(\mathrm{softmax}(\ell^{\bullet}_\theta(s_t))\big).
\end{aligned}
\end{equation}

\paragraph{\textbf{Exploration (server head)}} We mix $\varepsilon_t$-greedy exploration into the server branch and \emph{account for it in the likelihood}:

\begin{equation}
\begin{aligned}
\tilde{\pi}^{\mathrm{srv}}_\theta(a^{\mathrm{srv}}_t\!\mid\! s_t)
=(1-\varepsilon_t)\,\pi^{\mathrm{srv}}_\theta(a^{\mathrm{srv}}_t\!\mid\! s_t)
+\varepsilon_t\cdot \tfrac{1}{N},\\
\varepsilon_t=\max\!\Big(\varepsilon_{\min},\,\varepsilon_{\max}+\tfrac{t}{T_{\mathrm{dec}}}\big(\varepsilon_{\min}-\varepsilon_{\max}\big)\Big).
\end{aligned}
\end{equation}

The joint log-prob used for PPO is then
\begin{equation}
\begin{aligned}
\log \tilde{\pi}_\theta(a_t\mid s_t)=
\log \tilde{\pi}^{\mathrm{srv}}_\theta(a^{\mathrm{srv}}_t\mid s_t)+ \qquad\\ 
\log \pi^{\mathrm{w}}_\theta(a^{\mathrm{w}}_t\mid s_t)+
\log \pi^{\mathrm{g}}_\theta(a^{\mathrm{g}}_t\mid s_t).
\end{aligned}
\end{equation}

\paragraph{\textbf{Reward shaping}} Each scheduled block yields a scalar reward
\begin{equation}
\begin{aligned}
r_t=\alpha\ \underbrace{\tilde{p}_{\mathrm{acc}}}_{\text{accuracy prior}}
-\beta\,\underbrace{L_t}_{\text{latency (s)}}
-\gamma\,\underbrace{E_t}_{\text{energy (J)}} \\
-\delta\,\underbrace{\mathrm{Var}\!\big(\tfrac{1}{100}U^{(1..N)}_t\big)}_{\text{utilization imbalance}}
+\;b_t,
\end{aligned}
\end{equation}

where $\tilde{p}_{\mathrm{acc}}\in[0,1]$ is an empirical accuracy prior looked up from a width-combination table for the first $n$ segments (nearest-neighbor fallback) and the resulting correct or incorrect valuations for final segment; optionally we center it as $\tilde{p}_{\mathrm{acc}}\leftarrow \tilde{p}_{\mathrm{acc}}-\bar{p}_{\mathrm{top\text{-}1}}$ (zero-mean). $L_t$ is end-to-end latency for the block, $E_t=\bar{P}_t\cdot L_t$ uses the mean power across servers, and the imbalance term is the variance of normalized utilizations. $b_t$ is an optional bonus.

\paragraph{\textbf{Returns and advantages}} We use one-step returns and baseline:
\begin{equation}
\begin{aligned}
R_t \equiv r_t,\qquad A_t = R_t - V_{\theta_{\text{old}}}(s_t),\qquad
\hat{A}_t=\frac{A_t-\mu_A}{\sigma_A+\varepsilon}.
\end{aligned}
\end{equation}

\paragraph{\textbf{Clipped PPO objective}} With importance ratio
\begin{equation}
\begin{aligned}
\rho_t(\theta)=\exp\!\big(\log\tilde{\pi}_\theta(a_t\mid s_t)-\log\tilde{\pi}_{\theta_{\text{old}}}(a_t\mid s_t)\big),
\end{aligned}
\end{equation}

the clipped surrogate, value loss, and entropy regularizer are
\begin{equation}
\begin{aligned}
\mathcal{L}^{\mathrm{CLIP}}(\theta)=
\mathbb{E}\big[\min\big(\rho_t(\theta)\hat{A}_t,\; \qquad\\\mathrm{clip}(\rho_t(\theta),1-\epsilon,1+\epsilon)\hat{A}_t\big)\big],
\end{aligned}
\end{equation}

\begin{equation}
\begin{aligned}
\mathcal{L}^{V}(\theta)=\tfrac{1}{2}\,\mathbb{E}\big[(R_t - V_\theta(s_t))^2\big]
\end{aligned}
\end{equation}

\begin{equation}
\begin{aligned}
\mathcal{H}(\theta)=\mathbb{E}\big[H(\pi^{\mathrm{srv}}_\theta)+H(\pi^{\mathrm{w}}_\theta)+H(\pi^{\mathrm{g}}_\theta)\big].
\end{aligned}
\end{equation}

We minimize the total loss
\begin{equation}
\begin{aligned}
\mathcal{J}(\theta)= -\mathcal{L}^{\mathrm{CLIP}}(\theta)+c_v\,\mathcal{L}^{V}(\theta)-c_H\,\mathcal{H}(\theta),
\end{aligned}
\end{equation}

with clipping $\epsilon{=}0.2$, $c_v{=}0.5$, and entropy weight $c_H$ set by a hyperparameter. We run $K$ optimization epochs per update (here $K{=}3$) with gradient-norm clipping.



\section{Testing and Results}
\label{sec:results}

\subsection*{1. Datasets and Model}
We train and evaluate our Slimmable Model and Slim Scheduler setup on the CIFAR-100 data. The model for our scheduler is a slimmable SlimResNet partitioned into four sequential segments. Each segment supports width ratios $w\in\{1.00,\,0.75,\,0.50,\,0.25\}$. We employ Group Normalization instead of Batch Normalization to avoid cross-width statistics drift, following same values used in universal slimmable models. Though for learning rate scheduling we implement a cosine scheduler for increased model exploration as apposed to a linear scheduled learning rate. Before evaluating scheduling performance, we first verify the accuracy of the SlimResNet backbone across width configurations. Table~\ref{tab:uniform_slim_acc} reports Top-1 accuracy for uniformly slimmed networks utilizing width ratios $w\in\{1.00,\,0.75,\,0.50,\,0.25\}$, while Table~\ref{tab:mixed_slim_acc} lists results for four random mixed-width ratios sampled from a fixed seed. These results confirm that the SlimResNet backbone maintains strong accuracy under both uniform and mixed-width settings

\subsection*{2. Hardware Setup}
Experiments were run on a heterogeneous 3-GPU cluster with two NVIDIA RTX~2080 Ti GPUs and one NVIDIA GTX~980 Ti GPU. All devices used contained 64 gigabytes of memory. For communication between devices we utilized the University of California Irvine WLAN (Wi-Fi 5, 802.11ac).

\subsection*{3. Evaluation of Model Under Loads}
\begin{figure*}[!t]
  \centering
  \subfloat[Layer 1]{\includegraphics[width=0.24\textwidth]{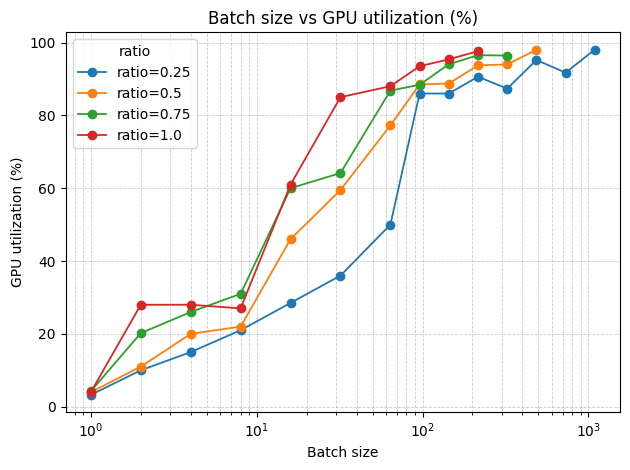}}
  \hfill
  \subfloat[Layer 2]{\includegraphics[width=0.24\textwidth]{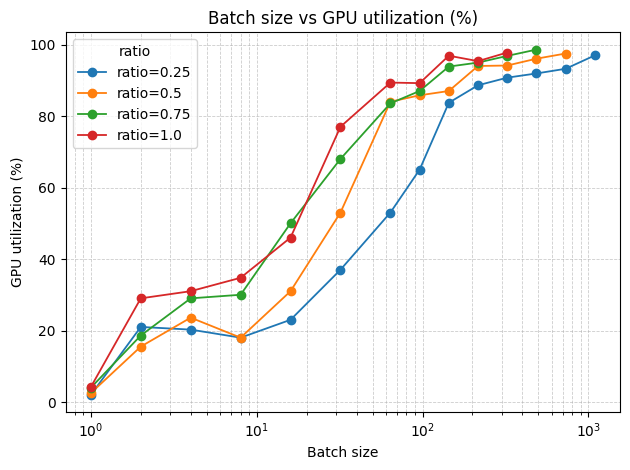}}
  \hfill
  \subfloat[Layer 3]{\includegraphics[width=0.24\textwidth]{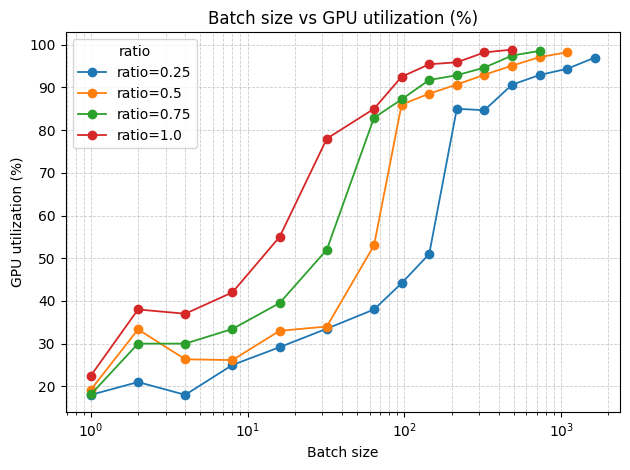}}
  \hfill
  \subfloat[Layer 4]{\includegraphics[width=0.24\textwidth]{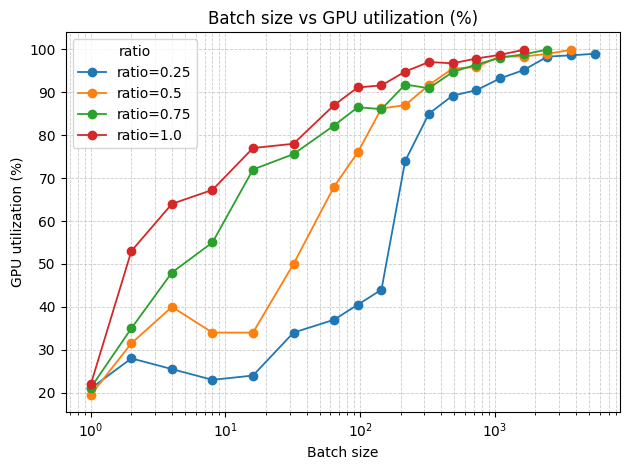}}
  \caption{GPU memory utilization vs.\ batch size for each segment on the RTX~2080~Ti.}
  \label{fig:cluster_util_batch}
\end{figure*}

To characterize how batching and slimming affect device-level efficiency, we evaluate a single RTX~2080~Ti GPU across varying batch sizes and width ratios. As shown in Figure~\ref{fig:cluster_util_batch}, GPU utilization increases steadily with batch size for all width ratios, with higher widths saturating memory and compute earlier. Narrower configurations (e.g., $0.25\times$ and $0.50\times$) maintain lower utilization at the same batch size, enabling smoother scaling before resource limits are reached.

\begin{figure*}[!t]
  \centering
  \subfloat[Layer 1]{\includegraphics[width=0.24\textwidth]{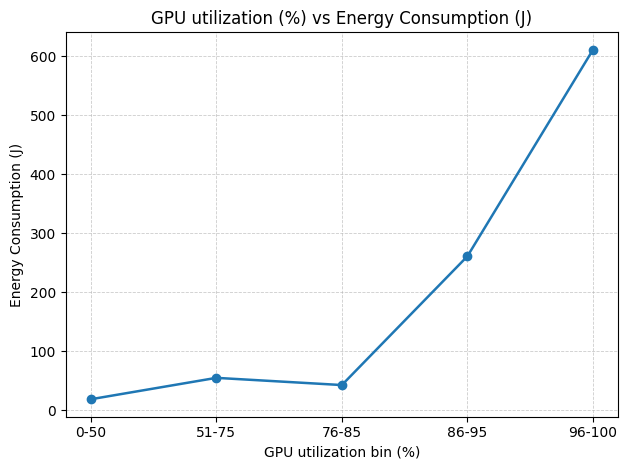}}
  \hfill
  \subfloat[Layer 2]{\includegraphics[width=0.24\textwidth]{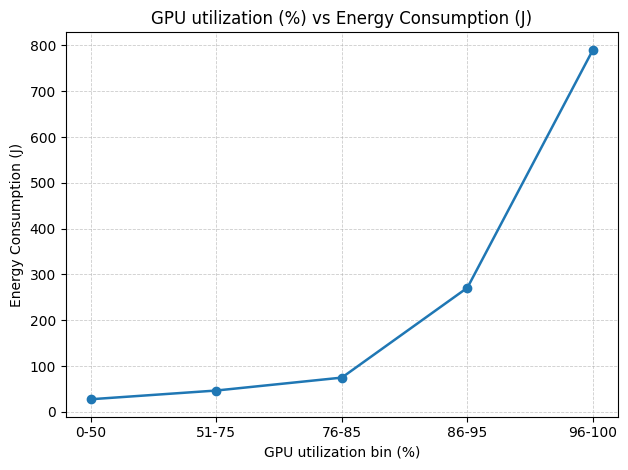}}
  \hfill
  \subfloat[Layer 3]{\includegraphics[width=0.24\textwidth]{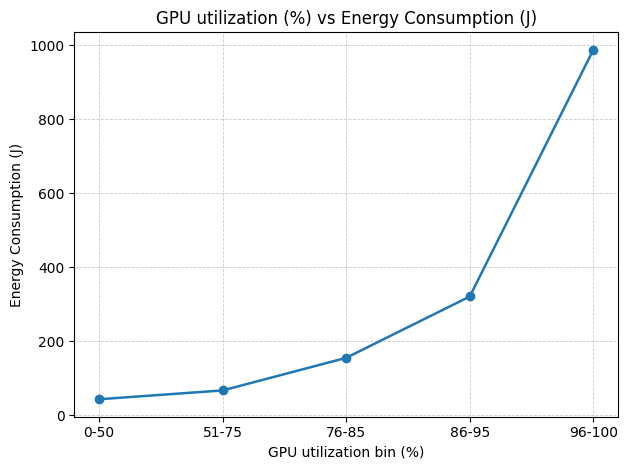}}
  \hfill
  \subfloat[Layer 4]{\includegraphics[width=0.24\textwidth]{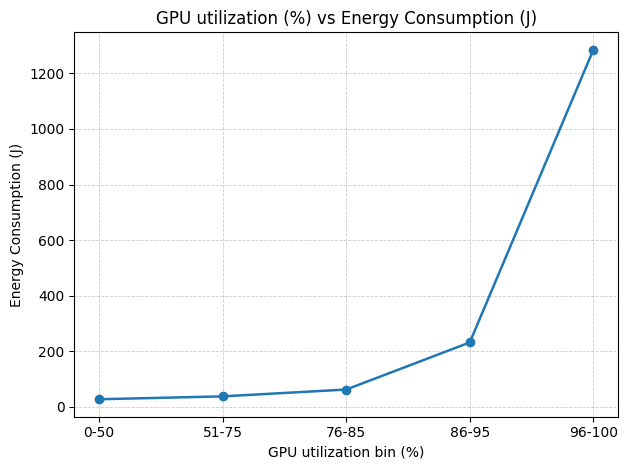}}
  \caption{Energy consumption vs.\ GPU utilization for each network on the RTX~2080~Ti.}
  \label{fig:cluster_latency_batch}
\end{figure*}

\begin{figure*}[!t]
  \centering
  \subfloat[Layer 1]{\includegraphics[width=0.24\textwidth]{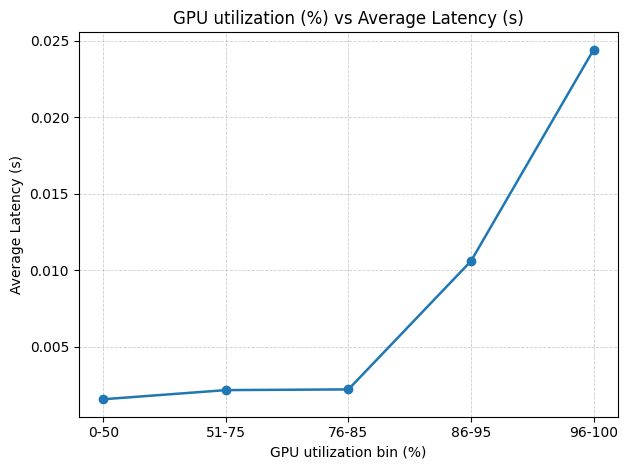}}
  \hfill
  \subfloat[Layer 2]{\includegraphics[width=0.24\textwidth]{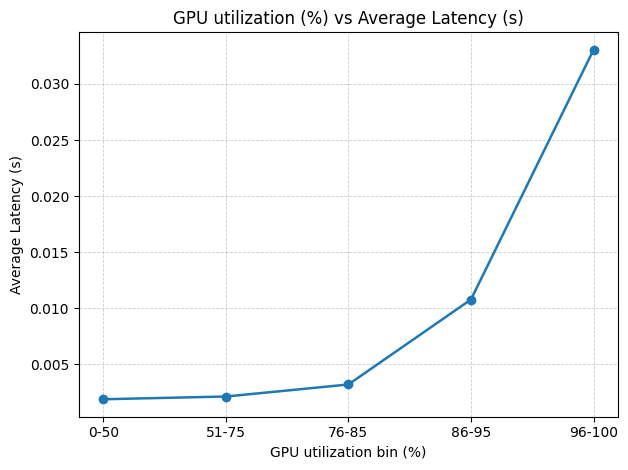}}
  \hfill
  \subfloat[Layer 3]{\includegraphics[width=0.24\textwidth]{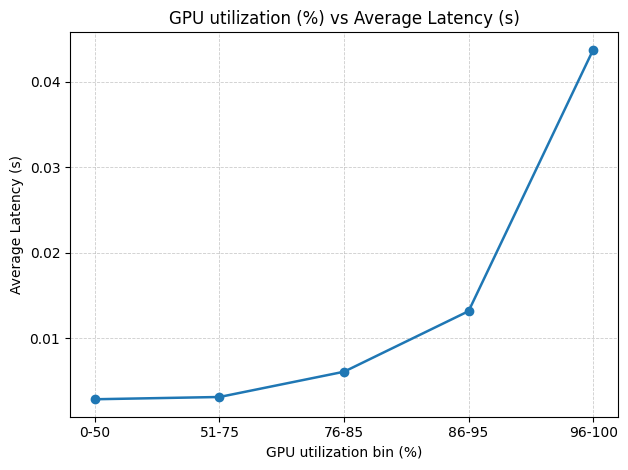}}
  \hfill
  \subfloat[Layer 4]{\includegraphics[width=0.24\textwidth]{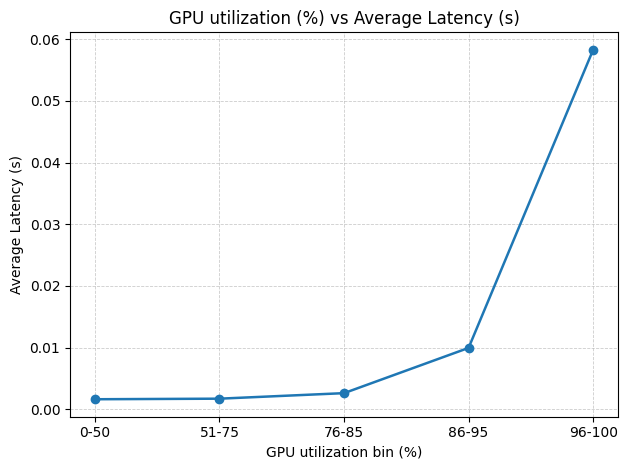}}
  \caption{Average latency vs.\ GPU utilization for each segment on the RTX~2080~Ti.}
  \label{fig:cluster_energy_batch}
\end{figure*}

Figures~\ref{fig:cluster_latency_batch} and~\ref{fig:cluster_energy_batch} reveal the downstream effects of this utilization growth. As utilization rises, both latency and energy consumption follow a near-linear trend up to roughly 90--95\% utilization. Beyond this threshold, the relationship becomes sharply nonlinear: small increases in utilization result in disproportionate spikes in latency and power draw. This saturation point reflects the practical limit of the 2080~Ti’s compute and memory bandwidth, where queueing delays and context-switch overheads dominate overall runtime.

These results establish the core relationship governing later experiments: larger batches drive higher GPU utilization, which in turn increases both latency and energy. The inflection near full utilization motivates the weighting of latency and energy terms ($\beta$ and $\gamma$) in the PPO reward function, since operating close to saturation yields diminishing returns in throughput while severely degrading efficiency.

\renewcommand{\arraystretch}{1.05}

\begin{table}[!t]
\centering
\caption{SlimResNet Top-1 accuracy under uniform width ratios (CIFAR-100)}
\label{tab:uniform_slim_acc}
\setlength{\tabcolsep}{5pt}
\resizebox{\columnwidth}{!}{
\begin{tabular}{lcccc}
\toprule
\textbf{Width Ratio} ($w_1{=}w_2{=}w_3{=}w_4$) & 0.25 & 0.50 & 0.75 & 1.00 \\
\midrule
\textbf{Top-1 Accuracy (\%)} & 70.30 & 72.99 & 74.93 & 76.43 \\
\bottomrule
\end{tabular}}
\end{table}

\begin{table}[!t]
\centering
\caption{SlimResNet Top-1 accuracy under randomized mixed-width ratios (CIFAR-100)}
\label{tab:mixed_slim_acc}
\setlength{\tabcolsep}{5pt}
\resizebox{\columnwidth}{!}{
\begin{tabular}{lc}
\toprule
\textbf{Width Ratio} $(w_1, w_2, w_3, w_4)$ & \textbf{Top-1 Accuracy (\%)} \\
\midrule
(1.00, 0.75, 0.50, 0.25) & 71.35 \\
(0.75, 1.00, 0.25, 0.50) & 72.33 \\
(0.50, 0.25, 1.00, 0.75) & 74.53 \\
(0.25, 0.50, 0.75, 1.00) & 75.33 \\
\bottomrule
\end{tabular}}
\end{table}

\subsection*{4. Results on a 3-GPU Cluster}
Tables~\ref{tab:baseline_results}, \ref{tab:ppo_overfit_results}, and~\ref{tab:ppo_average_results}
summarize performance on the heterogeneous 3-GPU cluster (2$\times$RTX~2080 Ti and 1$\times$GTX~980 Ti).
Table~\ref{tab:baseline_results} shows the greedy baseline using uniform random routing,
while Tables~\ref{tab:ppo_overfit_results} and~\ref{tab:ppo_average_results} correspond to two PPO\,+\,greedy
schedulers trained under different reward weightings.

The baseline configuration achieves stable accuracy (74.43\%) but suffers from high mean latency (8.98 s)
and energy consumption (1968 J), reflecting inefficient batching and underutilized cross-device coordination.
In contrast, the Slim Scheduler learns to minimize these costs by adaptively adjusting batch sizes
and slimming ratios according to load conditions.

When latency and energy penalties $(\beta,\sigma)$ are heavily weighted
(Table~\ref{tab:ppo_overfit_results}), the PPO policy converges toward using only the slimmest
($0.25\times$) configurations, yielding the same 70.3\% accuracy as the smallest model in isolation.
This configuration achieves the largest reductions in mean latency (-96.45\%) and energy (-97.31\%)
relative to the baseline, as the router consistently prioritizes low-cost inference paths and avoids
overloading any device. However, this aggressive optimization reduces overall throughput diversity,
and the network sacrifices accuracy for efficiency.

Relaxing the latency and energy weights to encourage balanced exploration
(Table~\ref{tab:ppo_average_results}) results in a policy that mixes wider models across servers,
recovering accuracy (75.26\%) and improving mean performance across runs.
Yet this comes at the expense of higher variance in both latency and energy
(standard deviations of 11.67 s and 2125 J, respectively), which reduces throughput stability.
This variance reflects the scheduler’s dynamic experimentation with different slimming ratios
to balance speed and accuracy under varying load conditions.

Overall, the Slim Scheduler exposes a clear trade-off surface:
strong optimization of latency and energy drives the policy toward uniformly slim models,
while more balanced weighting yields higher accuracy but greater runtime variability.
Although neither PPO configuration strictly outperforms the baseline in all metrics,
both demonstrate learned, resource-aware scheduling behavior that adapts effectively
to heterogeneous GPU capabilities.

\begin{table}[!t]
\centering
\caption{Baseline scheduler results on CIFAR-100 (3-GPU cluster)}
\label{tab:baseline_results}
\resizebox{250px}{!}{
\begin{tabular}{lccc}
\toprule
\textbf{Metric} & \textbf{Mean($\mu$)} & \textbf{Standard Deviation($\sigma$)} \\
\midrule
Accuracy (\%) & 74.43 \\
Latency (ms) & 8.979 & 7.302   \\
Energy (J) & 1967.94 & 1629.53   \\
GPU Var (\%) & 0.0433 & 0.0216  \\
\midrule
Image completion throughput & 250906 & \\
\bottomrule
\end{tabular}}
\end{table}

\begin{table}[!t]
\centering
\caption{PPO\,+\,Greedy scheduler results on CIFAR-100 (3-GPU cluster, overfit)}
\label{tab:ppo_overfit_results}
\resizebox{250px}{!}{
\begin{tabular}{lccc}
\toprule
\textbf{Metric} & \textbf{Mean($\mu$)} & \textbf{Standard Deviation($\sigma$)} \\
\midrule
Accuracy (\%) &70.30 \\
Latency (ms) & 0.318 & 0.755  \\
Energy (J) & 52.85 & 131.46  \\
GPU Var (\%) & 0.0633 & 0.0571  \\
\midrule
Image completion throughput & 420538 & \\
\bottomrule
\end{tabular}}
\end{table}

\begin{table}[!t]
\centering
\caption{PPO+Greedy scheduler results on CIFAR-100 (3-GPU cluster, averaged)}
\label{tab:ppo_average_results}
\resizebox{250px}{!}{
\begin{tabular}{lccc}
\toprule
\textbf{Metric} & \textbf{Mean($\mu$)} & \textbf{Standard Deviation($\sigma$)} \\
\midrule
Accuracy (\%) & 75.26\\
Latency (ms) & 6.100 & 11.673\\
Energy (J) & 1085.41 & 2125.62\\
GPU Var (\%) & 0.0815 & 0.0374\\
\midrule
Image completion throughput & 196947 & \\
\bottomrule
\end{tabular}}
\end{table}

\subsection*{5. Analysis of Results}
Across both the single-GPU and 3-GPU experiments, the results confirm that batch size, utilization, and energy form a tightly coupled feedback loop. Higher batch sizes increase GPU utilization, which improves short-term throughput but drives up both latency and power consumption—especially once utilization exceeds 95\%. The Slim Scheduler learns to exploit this behavior by reducing the operating point of each GPU just below the saturation threshold, minimizing total cost without explicit modeling of the device dynamics.

The two PPO configurations highlight opposing optimization extremes. When latency and energy penalties are dominant, the scheduler consistently selects the slimmest network configuration, yielding large efficiency gains but reduced accuracy and throughput diversity. Conversely, with relaxed penalty weights, the scheduler recovers accuracy through broader model utilization at the cost of higher variance and inconsistent latency. These findings illustrate the inherent trade-off between efficiency and stability: tighter optimization favors predictable but narrower operating regimes, while balanced objectives introduce controlled variability that can better utilize heterogeneous resources.

Overall, the learned scheduler generalizes across devices and resource levels, demonstrating that reinforcement learning can capture meaningful system dynamics without direct supervision. The results suggest that future work should explore adaptive reward scaling or uncertainty-aware policies to maintain efficiency while improving predictability in multi-device inference.

\section{Conclusion}
In summary, the proposed Slim Scheduler dynamically adapts inference workloads across heterogeneous GPUs, achieving significant reductions in mean latency and energy consumption. These gains come with a measured tradeoff of higher variance in latency and energy, caused by the scheduler’s adaptive adjustments of slimming and batching configurations. Overall, the results demonstrate that controlled runtime variability can yield large efficiency improvements, providing a scalable foundation for resource-aware, multi-device neural network inference.

\bibliographystyle{IEEEtran}
\bibliography{references}

\end{document}